\documentclass[conference]{IEEEtran}
\usepackage{booktabs} 
\usepackage{graphicx}
\usepackage{subcaption}
\usepackage{multirow}
\usepackage{algorithm2e}
\usepackage{tabularx}
\usepackage{tabulary}
\usepackage{float}
\usepackage{amsmath}
\usepackage{cleveref}
\usepackage{makecell}
\usepackage{color}
\renewcommand\baselinestretch{0.95}

\ifCLASSINFOpdf
\else
\fi

\begin{document}
%
\title{Mission-Aware Spatio-Temporal Deep Learning Model for UAS Instantaneous Density Prediction}

\author{\IEEEauthorblockN{Ziyi Zhao$^{1}$, Zhao Jin$^{1}$, Wentian Bai$^{1}$, Wentan Bai$^{1}$, Carlos Caicedo$^{2}$, M. Cenk Gursoy$^{1}$, Qinru Qiu$^{1}$}
\IEEEauthorblockA{$^{1}$Department of Engineering and Computer Science, Syracuse University, Syracuse, New York, USA\\
$^{2}$School of Information Studies, Syracuse University, Syracuse, New York, USA\\
Email: { $^1$\{zzhao37, zjin04, wbai02, wbai03, mcgursoy, qiqiu\}@syr.edu,$^2$\@ccaicedo@syr.edu} }
}


%


\maketitle

\begin{abstract}
The number of daily sUAS operations in uncontrolled low altitude airspace is expected to reach into the millions in a few years. Therefore, UAS density prediction has become an emerging and challenging problem. In this paper, a deep learning-based UAS instantaneous density prediction model is presented. The model takes two types of data as input: 1) the historical density generated from the historical data, and 2) the future sUAS mission information. The architecture of our model contains four components: Historical Density Formulation module, UAS Mission Translation module, Mission Feature Extraction module, and Density Map Projection module. The training and testing data are generated by a python based simulator which is inspired by the multi-agent air traffic resource usage simulator (MATRUS) framework. The quality of prediction is measured by the correlation score and the Area Under the Receiver Operating Characteristics (AUROC) between the predicted value and simulated value. The experimental results demonstrate outstanding performance of the deep learning-based UAS density predictor. Compared to the baseline models, for simplified traffic scenario where no-fly zones and safe distance among sUASs are not considered, our model improves the prediction accuracy by more than 15.2\% and its correlation score reaches 0.947. In a more realistic scenario, where the no-fly zone avoidance and the safe distance among sUASs are maintained using A* routing algorithm, our model can still achieve 0.823 correlation score. Meanwhile, the AUROC can reach 0.951 for the hot spot prediction. 

\end{abstract}


%
\IEEEpeerreviewmaketitle

\section{Introduction}\label{sec:introduction}
Recently, many companies, such as DJI, Lockheed Martin and Amazon, devote themselves to develop small Unmanned Aircraft Systems (sUAS). Complicated and high density UAS traffic imposes significant burden on air traffic management, city planning and communication resource allocation. Under this environment, the following critical questions are usually asked: Given the list of scheduled launches in an area, do we know in advance whether a feasible route in terms of air space safety and energy efficiency can be found for a specific mission at a specific time? Do we need to delay the launch of some sUAS in advance to accommodate a mission with higher priority scheduled at a specific time? Answering such questions and being able to predict the traffic distribution ahead of time will provide an opportunity for more efficient planning and control. 

UAS density prediction is a critical and challenging problem in the Unmanned Aircraft System Traffic Management (UTM) system. Most existing studies focus on simulation-based approaches. Although accurate, they usually take a long time to deliver results. Neural networks have been used to predict the traffic density. However, most such studies require the sampling of the traffic density from the past data and predict the future density using past density information. These models assume a static environment. For example, the source (i.e. the location where the sUAS enters the air space) and sink (i.e. the location where the sUAS leaves the air space) of the traffic flow are assumed to remain the same, and air space constraints, such as no-fly zones, are fixed. Based on these assumptions, the traffic in the future will exhibit similar pattern as the traffic in the past, and can be predicted from the historical data. A constant environment may be reasonable for road traffic, however, the operational environment of sUAS features higher dynamics and flexibility. For instance, the no-fly zones may change due to construction or special activities/events, launching or landing zones may be added or removed. The model based on historical data will become obsolete as soon as the environment changes. New data must be collected and a new model needs to be trained, which can take days or weeks. Furthermore, most of the existing models consider traffic distribution as a stationary process, and focus on predicting the steady states. For resource provisioning or safety assurance, we need to know not only the steady state traffic but also the worst case traffic. Hence the ability to predict the transient behavior of air traffic distribution is highly desirable. There are a few works that utilize the long short-term memory (LSTM) model to predict future traffic based on recent traffic activities, however, their prediction horizon is very limited. Accurate prediction cannot be made beyond 4 or 5 timestamps. 

In this paper, a deep learning-based prediction model is presented for semi-transient traffic density distribution prediction. The model takes the air space environment and the pre-scheduled launch list in the next T time units as the inputs, and predict the average traffic density of traffic distribution in this air space during time [T-$\delta$, T]. The parameter T controls the prediction horizon and by reducing the value of parameter $\delta$ , the focus of the model changes from the long-term average behavior to transient behavior of the traffic. By taking the flight environment and detailed launch information as part of the inputs, the model is specific only to the type of trajectory planning algorithms. It can be generalized to different air space environment as long the trajectory of each UAS is routed using the same algorithm. It will have no “down time” after the map or the launching/landing zone has changed. 

The model has high prediction accuracy. Compared with other existing methods, our model can achieve a correlation score of 0.947 and can improve the prediction accuracy by up to 15.2\%. In a realistic traffic scenario, where no-fly zones avoidance and safe distance between sUASs are considered by planning the trajectory using A* routing algorithm \cite{zhao2019temporal}, our model can still achieve a correlation score of 0.823.
The following summarizes the major contributions of our work:

\begin{itemize}
\item A novel UAS traffic density prediction model is developed that captures the information from the historical data and the pre-scheduled sUAS launch list.

\item A novel input representation of the future sUAS mission information is proposed. The pre-scheduled missions are categorized into 3 types according to their launching times. Our model is designed to extract features from all type of inputs simultaneously. The learnable parameters are introduced to adjust the degrees affected by different types of features. 

\item Compared to the baseline models, our model improves the prediction accuracy by up to 15.2\%. When doing hot spot prediction, our model can achieve an AUROC score of 0.951. Meanwhile, the qualitative results demonstrate that our model can accurately predict the hot spot in the future traffic map. 
\end{itemize}


\section{Related Works} \label{sec:related_works}

Over the past decade, the unmanned aerial vehicles have played an increasingly essential role in many areas \cite{puri2005survey} \cite{gupta2013review} \cite{liu2014review} \cite{nex2014uav}. With the rise in the popularity of sUAS, many notable issues telated to UAS traffic management have been discovered. However, most of them explored the novel applications for the single sUAS or formulated the UAS management policies. The study of the UAS cluster behaviors such as forecasting of the UAS traffic density has generally not been addressed. In our investigation, the density forecasting approaches can be categorized into simulation based method and deep-learning based method. In this section, we will analyze the pros and cons of  recent works in these two categories.

Many existing works study issues such as sUAS navigation, obstacle avoidance or UAS traffic management, by developing a corresponding simulator with fair time complexity. In \cite{odelga2016obstacle}, the authors presented an indoor algorithm to navigate single sUAS to avoid collisions. \cite{thanh2018completion} proposed a solution to avoid collisions in a static environment by importing geometrical constraints. Other single sUAS classical approaches applied rapidly-exploring random trees \cite{lavalle1998rapidly} and Voronoi graphs \cite{bortoff2000path} \cite{tisdale2009autonomous}. 
Multiple  sUAS  trajectory simulation has  been studied as a multi-agent cooperative system and solved in a rolling  horizon  approach  using  dynamic  programming \cite{beard2003multiple} or mixed integer linear programming \cite{song2016rolling}. Other strategies \cite{zhao2019temporal} \cite{jin2020simulation} involved real-time routing algorithms with communication and airspace safety considerations. Recently, a very strict and rigid airspace structure to handle dense operation in the urban low altitude environment was suggested by the work on Unmanned Air craft System (UAS) Traffic Management (UTM) at NASA in \cite{jang2017concepts}. The authors explored UAS operations in non-segregated air space and managed the risk of mid-air collision to a level deemed acceptable to regulators. In the paper, the airspace was divided into  multiple layers, and the layers were further divided into orthogonal sky lanes. There are no current works that solve the traffic prediction problem from a big picture perspective within a small running time.

In this work, instead of developing a simulator, we utilize deep learning for UAS traffic density prediction. The deep learning based approach has shown outstanding success in many application domains \cite{zhao2018learning} \cite{fang2019event} \cite{fang2019general} \cite{ylu03}. Similar multi-agent works are addressed in other fields such as pedestrian density prediction and autonomous driving \cite{ylu06} \cite{ylu02} . In \cite{zhao2019multi}, the author proposed a LSTM based scene-aware model to predict trajectories for autonomous driving. However, the prediction errors grew exponentially as the time horizon increased. Another work addressed pedestrian traffic flow prediction by fusing historical information, but the prediction is limited by historical data regardless of upcoming event information. Existing single agent trajectory prediction works concentrated on the behavior of a single sUAS and the impact of environment conditions, without any sUAS cluster consideration. For example, \cite{shi2018lstm} proposed a LSTM-based flight trajectory model with weather considerations taken into account. \cite{li2019autonomous}\cite{eslamiatautonomous} aimed to solve environment navigation problems, and developed a reinforcement learning model to plan energy efficient waypoint in a static environment.

Compared to the existing work, our model has the ability to learn and extract the information from the historical data and the pre-scheduled sUAS mission launch list. And the error is restrained strictly via dynamic feature extraction. By adopting a novel channel segmentation approach, our mission feature extraction module can learn the density features accurately.

\section{Methods}\label{sec:methods}

\begin{figure*}[htb]
\centering
\includegraphics[width=0.75\textwidth]{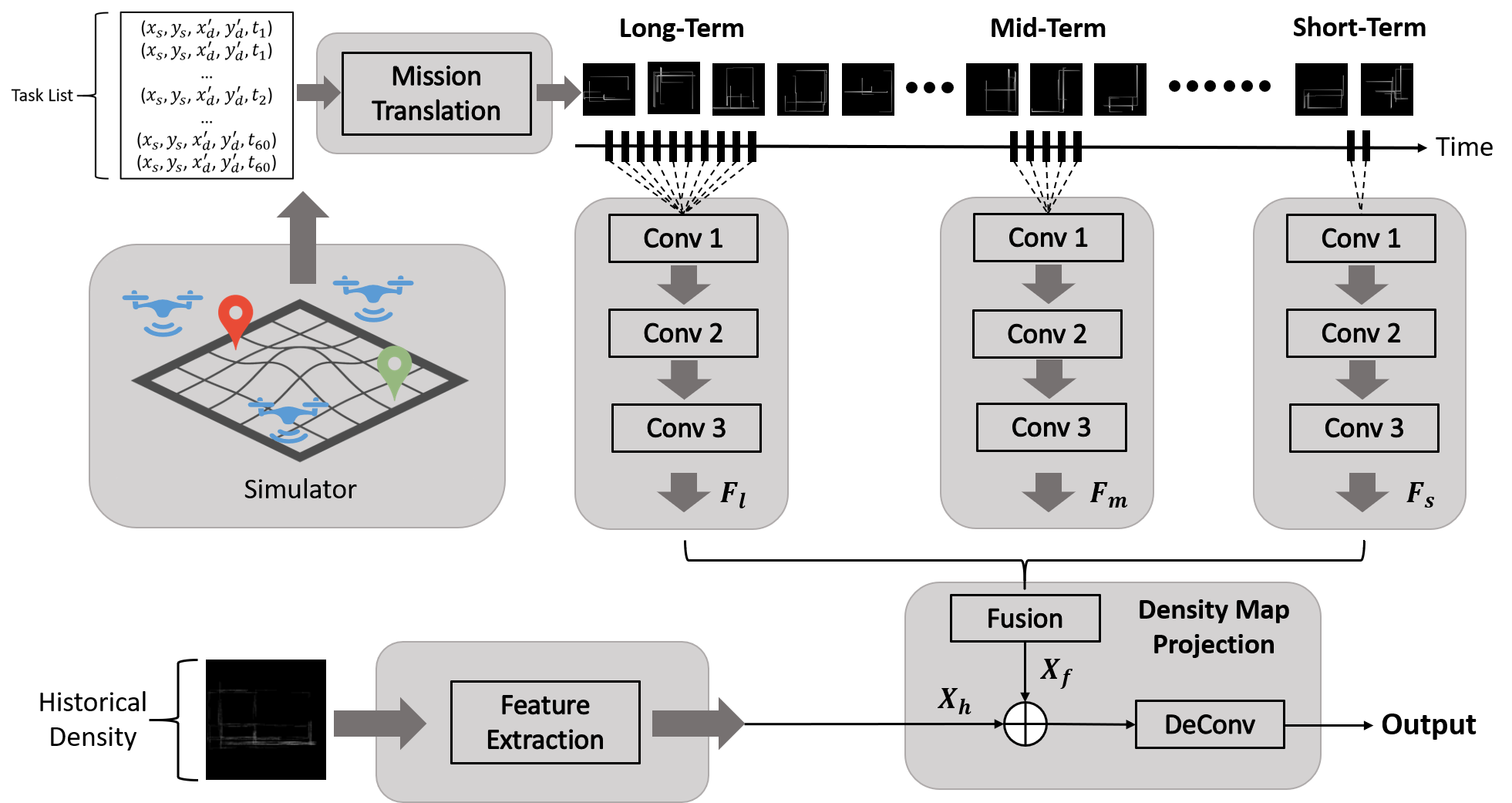}
\caption{Mission-Aware Spatio-Temporal Model Architecture}
\label{fig:architecture}
\vspace{-1em}
\end{figure*}

Our density prediction model is an end-to-end model where each module is fully differentiable. The mean square error (MSE)  loss is calculated by measuring the difference between the predicted density map and the labeled density map. The architecture of our model contains four components: Historical Density Formulation module, UAS Mission Translation module, Mission Feature Extraction module, and Density Map Projection module. The model structure is depicted in Figure \ref{fig:architecture}. In this section, each component of the model will be elaborated.

\subsection{Historical Density Formulation}\label{sec:historical}

The historical density describes the pre-existing air space environment. The size of the historical density map is \begin{math} 100*100 \end{math} grid units which is the same as the simulation environment. The value at each pixel is between 0 and 1 and represents the average density in the past \begin{math} m \end{math} simulation cycles. The value of \begin{math} m \end{math} is set to be 10 in this paper. The historical density will also be called ``initial density'' in this paper. Given the historical density, we employ a convolutional neural network (CNN) to extract the relevant features. The model is composed of 3 convolution layers, 2 pooling layers and the ReLU activation layers. Finally, the \begin{math} C*32*32 \end{math} feature maps are obtained from the feature extractor, where \begin{math} C \end{math} is the number of feature channels. The feature extracted from the historical density is denoted as \begin{math} X_h \end{math}.

\subsection{UAS Mission Translation}\label{sec:mission}
This module is responsible for translating the UAS missions to the image representation. First, the UAS future missions are summarized into a mission list. The dimension of the mission list is \begin{math} n*5 \end{math} where \begin{math} n \end{math} is the number of missions in the future. Each mission is defined by a \begin{math} 5 \end{math} dimensional vector: \begin{math} \{X_s, Y_s, X^\prime_d, Y^\prime_d, T_n\} \end{math}. The \begin{math} \{X_s, Y_s\} \end{math} and the \begin{math} \{X^\prime_d, Y^\prime_d\} \end{math} represent the launching and landing locations of the mission. The launching time is indicated by \begin{math} T_n \end{math}. Given the mission list, the model will first cluster the missions into different groups based on the mission launching time. The mission translation module follows in the same manner as the BFS algorithm to map the trajectory into a \begin{math} 2D \end{math} map. From each Origin-Destination (O-D) pair, we draw a shortest path from the launching location to the landing location. For each individual mission, we assume that the horizontal direction movement has a higher priority than the vertical direction movement. The movement priority is the same as the MATRUS simulator \cite{zhao2019simulation}. After the UAS mission translation procedure as mentioned above, a \begin{math} K \end{math} channel output can be obtained. Each channel lumps the trajectory information of the sUAS that will be launched at the same simulation cycle. \begin{math} K \end{math} is set to be 60 in this paper. 

\begin{figure}[!htbp]
\centering
\begin{subfigure}{0.3\linewidth}
  \includegraphics[width=1.1in]{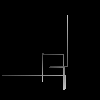}
  \caption{Translation 1}
  \label{fig:sample_2}
\end{subfigure}%
\hspace{0.15\linewidth}
\begin{subfigure}{0.3\linewidth}
  \includegraphics[width=1.1in]{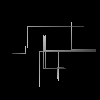}
  \caption{Translation 2}
  \label{fig:sample_3}
\end{subfigure}
\caption{UAS Mission Translation Examples}
\label{fig:sample}
\vspace{-1em}
\end{figure}

Moreover, we introduce a novel sUAS trajectory representation approach, which we call as ``Flow''. The ``Flow'' input representation uses an ascending sequence to represent the sUAS movement from the launching location to the landing location. Therefore, in the \begin{math} 2D \end{math} map, the waypoints near the landing location are brighter than the waypoints near the launching location. If one location is occupied by more than one sUAS, we use the mean  of all the overlapped values to represent this location. By using this input representation, the model can distinguish launching and landing locations. In addition, the order of the sUAS movement is also specified. 
Figure~\ref{fig:sample} shows two mission translation examples.


\subsection{Mission Feature Extraction}\label{sec:t_net}
The translated UAS missions are fed into the mission feature extraction module. This module is responsible for learning the density features from the pre-scheduled missions. Inspired by \cite{zhang2017deep}, we develop a novel channel segmentation model. First, according to the mission launching time, the translated missions are categorized into 3 groups: long-term, mid-term and  short-term. In our case, the long-term group contains the launching missions from cycle 1 to cycle 30. The mid-term group involves the launching missions from cycle 31 to cycle 50. The rest of the launching missions, cycle 51 to cycle 60, belong to the short-term group. Then, three types of models with different number of input channels are employed to extract the features from the inputs. The number of input channels for long-term, mid-term and short-term models are 10, 5 and 2, respectively. Each individual model has the same structure but the weight will be updated independently. The intuition of the model architecture design is that the mission whose launching time is close to the end should have a larger impact on the final density. The type of convolution (2D or 3D) operation we applied in this module will be discussed in Section~\ref{sec:2d_3d}. Then, three types of features (long-term feature, mid-term feature and short-term feature) can be obtained, which are denoted as \begin{math} \{F_l, F_m, F_s\} \end{math}. In the fusion module, the learnable parameters are introduced to adjust the degrees affected by different features. Therefore, the mission from different times will contribute accordingly to the final density. The fusion equation is defined as follows:
\begin{equation}
      X_f = \{W_1 * F_{s\_1} + ... W_k * F_{m\_1} + ... + W_{n} * F_{l\_n}\}  
\end{equation}
where \begin{math} W \end{math} denotes the learnable parameters. The output density feature is denoted as \begin{math} X_f \end{math}. And, the \begin{math} F_s \end{math}, \begin{math} F_m \end{math} and \begin{math} F_l \end{math} are the features extracted from short-term input, mid-term input and long-term input, respectively.

Consequently, a \begin{math} C*32*32 \end{math} feature map is obtained from this module, where \begin{math} C \end{math} stands for the number of feature channels.

\subsection{Density Map Projection}\label{sec:t_net}
Finally, two features \begin{math} \{X_h, X_f\} \end{math} are concatenated together to construct a fused density feature representation. Then, we apply a de-convolution module to project the density feature into a \begin{math} 2D \end{math} density map that has the same width and height as the whole simulation environment. The de-convolution module is composed of 4 2D-transpose layers, batch normalization layers and the ReLU activation layers. The value at each location stands for the average density at the given prediction time \begin{math} T_1 \end{math}. In this paper, \begin{math} T_1 \end{math} is set to be 10.

\section{Experiments}\label{sec:experiments}

\subsection{Data Generation}\label{sec:data}

Inspired by the MATRUS framework \cite{zhao2019simulation}, we implement a python based sUAS flight simulator. For each traffic scenario tested in this paper, we ran the simulator to generate 3000 samples.  All the data sets are divided into two subsets: training and testing. The split ratio is \begin{math} 90:10 \end{math}. For each sample, the simulator randomly generates 5 launching areas and 5 landing areas on a 100*100 grid environment. Each area has the size 3*3. Any grid in this area can be considered as the launching location. The minimum distance between any two areas is 5 cells. For each launching area, the simulator uses uniform distribution to randomly generate a float number as the launch probability. At every simulation cycle, the simulator randomly selects 15 launching locations from all launching areas. For each selected location, the simulator randomly decides whether a mission should be launched from current location at current cycle based on the launch probability.

\begin{figure}[!htbp]
\centering
\includegraphics[width=0.4\textwidth]{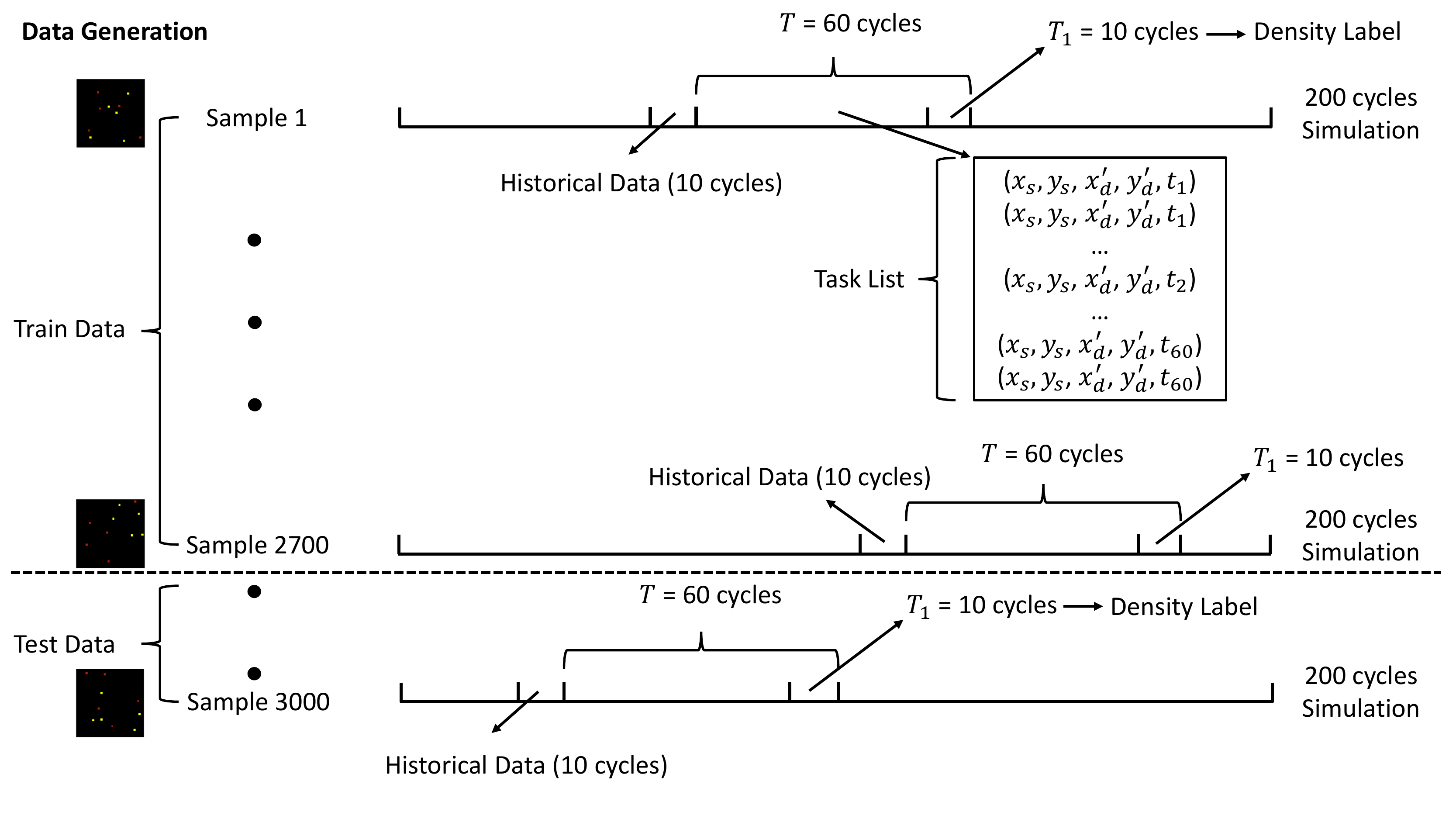}
\caption{Data Generation Procedure}
\label{fig:data_generation}
\vspace{-1em}
\end{figure}

For each sample, the simulation time horizon is defined as \begin{math} T \end{math}. In this paper, \begin{math} T \end{math} is set to be 60 simulation cycles, and each cycle lumps sUAS launching information in 10 seconds. The time period that generates the density map will be described as \begin{math}T_1 \end{math}. The data generation procedure is depicted in Figure~\ref{fig:data_generation}.

\subsection{Evaluation Metrics}\label{sec:metrics}
Two metircs are used to measure the quality of the prediction.

(1) Correlation: Correlation is calculated between the simulated traffic density \begin{math}(Y)\end{math}, which is considered as the ground truth, and the predicted traffic density \begin{math}(\hat{Y})\end{math}. In our experiments, it shows whether and how strongly the predicted and labeled variables are related. The equation of the Correlation is as follows:
\vspace{-0.1cm}
\begin{equation}
     \rho(X,Y) = \frac{Cov(X,Y)}{\sigma_{X}\sigma_{Y}} 
\end{equation}
\vspace{-0.1cm}
where $\rho(X,Y)$ is Pearson's correlation coefficient of X and Y, $\sigma_{X}$ and $\sigma_{Y}$ are the standard deviation of \begin{math} X \end{math} and \begin{math} Y\end{math}, respectively. \begin{math} Cov(X,Y) \end{math} is the covariance of variables \begin{math} X \end{math} and \begin{math} Y\end{math}, which can be calculated by the following equation: 
\vspace{-0.1cm}
\begin{equation}
     Cov(X,Y) = E[(X-E[X])(Y-E[Y])]
\end{equation}
\vspace{-0.1cm}
where \begin{math} E[.] \end{math} denotes the expected value.

(2) Area Under the Receiver Operating Characteristics (AUROC): The ROC curve is plotted by mapping the True Positive Rate (TPR) against the False Positive Rate (FPR) with different thresholds. Given a ROC curve, the AUROC evaluates the performance of the model by distinguishing between classes. The higher the AUROC score of a model, the better the performance is. For an uninformative model, the AUROC is close to 0.5. The maximum AUROC is 1.

\subsection{Comparison Models}\label{sec:models}
To the best of our knowledge, there is no prior work considers the exactly same application as this paper. For comparison, we selected some existing models that are potentially promising for traffic prediction, and re-trained them using our data set. We also compared with some modified version of our own model to show the effectiveness of certain design decisions of our model. The following five models are tested and compared. 
\begin{itemize}

\item Vanilla CNN (VCNN): This is a typical CNN based encoder-decoder model. The model assumes that, the location and action probability of each launching/landing area is static and can be represented in a 2D map. It tries to learn the relation between traffic density and the 2D map, and makes prediction based on the static information.

\item Vanilla LSTM (VLSTM): This is a typical LSTM based encoder-decoder model. It takes the \begin{math}T\end{math} cycles scheduled launching information and predict the density at the \begin{math}T+1\end{math} cycle. Because the traffic density of cycle \begin{math}(t+1)\end{math} is determined by the density at cycle \begin{math}t\end{math} and the current launching information, it was expected that such temporal dependency can be captured by an LSTM model.

\item RouteNet \cite{xie2018routenet}: The RouteNet model is encouraged by the Fully Convolutional Network (FCN) architecture that predicts the congestion in VLSI placement and route. And the FCN allows input to be any size and produces an output with exactly the same size as input, indicating the density (or hotspot) at any location. 

\item Segmented Channel: This is our model discussed in this paper. The inputs are categorized into 3 groups. Then, the designated models are assigned to each group for extracting the features.  

\item All Channel: This model has the similar structure as our model except that there is no input channel segmentation. The model treats all the missions from different launching time as the same. 

\end{itemize}


\section{Results}\label{sec:results}

\subsection{Predicted Density Accuracy Improvement}\label{sec:accuracy}
In the first experiment, we compare the accuracy of the density prediction between our model and other baseline models from Section~\ref{sec:models}. The model which extracts the trajectory features from all input channels is denoted as ``\begin{math} channel_{all} \end{math}''. Our presented channel segmentation model is denoted as ``\begin{math} channel_{seg} \end{math}''. Because the prediction model assumes a non-empty air space, we are interested to know how close the initial traffic density resembles the density at the target time of prediction. The column ``\begin{math} init \end{math}'' gives the correlation between the initial density and the label density.


\begin{table}[!htbp]

\begin{center}
{\caption{The Correlation Score of the Density Prediction}\label{tab:table_accuracy}}
\begin{tabular}{lccccccc}
\hline
\rule{0pt}{12pt}
Init & VCNN & VLSTM & RouteNet & $Channel_{all}$ & $Channel_{seg}$
\\
\hline
\\[-6pt]
0.822 & 0.863 & 0.803 & 0.889 & 0.944 & \textbf{0.947}
\\
\hline
\\[-6pt]
\end{tabular}
\end{center}
\vspace{-1em}
\end{table}


Table~\ref{tab:table_accuracy} shows the correlation score for all the models. The LSTM model has the worst performance among all the models. The correlation score of the ``\begin{math} VLSTM \end{math}'' model is even lower than the ``\begin{math} init \end{math}'' correlation score. One reason for this is that the 60 cycle prediction period is too long for the ``\begin{math} VLSTM \end{math}'' model. The error will accumulate and propagate from the first cycle to the last cycle. The ``\begin{math} VCNN \end{math}'' model improves the correlation score by 5.0\%. However, ignoring the information of exactly when and where each sUAS is going to be launched and where it is heading from now to the end of prediction window makes the prediction much less specific. Therefore, the ``\begin{math} VCNN \end{math}'' model cannot achieve a higher accuracy. In the ``\begin{math} RouteNet \end{math}'' model, each scheduled mission will be marked by a bounding box between the Origin-Destination (O-D) pair. This approach gives the model a more forthright indication of  each mission and the relation between the launching and the landing locations. Consequently, the ``\begin{math} RouteNet \end{math}'' model improves the correlation score by 8.2\%. Finally, our presented model, ``\begin{math} channel_{all} \end{math}'' and ``\begin{math} channel_{seg} \end{math}'', outperforms all other models. Compared to the initial traffic density, the predicted density of these models clearly resembles the actual density at the end of prediction window better, with 14.8\% and 15.2\% improvement of the correlation score, respectively.

Between the two model architectures that we proposed, the ``\begin{math} channel_{seg} \end{math}'' model can achieve higher correlation score than the ``\begin{math} channel_{all} \end{math}'' model. However, the difference is very marginal. In the next, we will show that using segmented channel is important under the scenario when the process of UAS launching is non-stationary. 

\subsection{The Impact of the Initial Density}\label{sec:init}

In the second experiment, we study how the initial density affects the density prediction and evaluate the robustness of two model architecture designs. Two test scenarios have been designed:

\begin{itemize}

\item Without Training (w/o training): We follow the same training procedure in Section~\ref{sec:accuracy}. However, during the testing, the initial density is not provided. Instead an all-black image (i.e. an empty air space) is provided.

\item With Training (w training): In the training phase, the initial density is also replaced by the all black image.

\end{itemize}

\begin{table}[!htbp]
\begin{center}
\caption{The Impact of the Initail State}\label{tab:init}
\begin{tabular}{lccc}
\hline
\rule{0pt}{12pt}
 & w/o training & w training
\\
\hline
\\[-6pt]
\quad \begin{math} init \end{math} & 0.822 & 0.822\\
\quad \begin{math} channel_{all} \end{math} & 0.885 (+7.7\%) & 0.894 (+8.8\%) \\
\quad \begin{math} channel_{seg} \end{math} & 0.913 (+11.1\%) & 0.924 (+12.4\%) 
\\
\hline
\\[-6pt]
\end{tabular}
\end{center}
\vspace{-1em}
\end{table}


The results in Table~\ref{tab:init} show that the model performance is affected by the historical density. Nonetheless, compared to the ``\begin{math} channel_{all} \end{math}'' model, the ``\begin{math} channel_{seg} \end{math}'' model is more robust. In Section~\ref{sec:accuracy}, the correlation scores are 0.944 and 0.947 respectively for ``\begin{math} channel_{all} \end{math}'' model and the ``\begin{math} channel_{seg} \end{math}'' model. Without the training, the correlation score drops to 0.895 and 0.913 respectively for ``\begin{math} channel_{all} \end{math}'' model and the ``\begin{math} channel_{seg} \end{math}'' model. This proves that the ``\begin{math} channel_{seg} \end{math}'' model can extract more meaningful features from the pre-scheduled missions. Even with the training, two models can only achieve correlation scores of 0.894 and 0.924. This proves that the historical density map can help the model improve the prediction accuracy. This phenomenon shows that, compared to the ``\begin{math} channel_{seg} \end{math}'' model, the ``\begin{math} channel_{all} \end{math}'' model relies more on the historical density.

\subsection{The Model Sensitivity to Missions}\label{sec:2d_3d}

In the third experiment, we study the model's sensitivity to missions. In our assumption, the most recent missions should have a larger contribution to the predicted density than those that took place earlier in time. And the model should be able to capture the features from the non-stationary missions. In order to further analyze this conjecture, we test different model architectures and design a specific experiment. In the model architecture design, we test both 2D convolution operation and 3D convolution operation to be the feature extractor backbone. For the experiment, we use the normal mission list as the input in the model training phase. The normal mission list means that all the 60 cycles have the launching missions. However, in the testing phase, we remove the missions from either the first 30 cycles or last 30 cycles. Therefore, the experiments are broken down into 2 scenarios:

\begin{itemize}

\item No Task Before 30 (NTB\_30): No new launching mission from cycle 1 to cycle 30.

\item No Task After 30 (NTA\_30): No new launching mission from cycle 31 to cycle 60.

\end{itemize}

\begin{table}[!htbp]
\begin{center}
\caption{Experiments comparing 2D and 3D convolution}\label{tab:2d_3d}
\begin{tabular}{lccc}
\hline
\rule{0pt}{12pt}
Scenario (3D) & NTB\_30 & NTA\_30
\\
\hline
\\[-6pt]
\quad \begin{math} init \end{math} & 0.699 & 0.647\\
\quad \begin{math} channel_{all} \end{math} & 0.554 (-21.3\%)  & 0.532 (-17.8\%)\\
\quad \begin{math} channel_{seg} \end{math} & 0.712 (+1.9\%) & 0.678 (+4.8\%)
\\
\hline
\\[-6pt]
\end{tabular}

\smallskip

\begin{tabular}{lccc}
\hline
\rule{0pt}{12pt}
Scenario (2D) & NTB\_30 & NTA\_30
\\
\hline
\\[-6pt]
\quad \begin{math} init \end{math} & 0.699 & 0.647\\
\quad \begin{math} channel_{all} \end{math} & 0.765 (+9.4\%)  & 0.751 (+16.1\%)\\
\quad \begin{math} channel_{seg} \end{math} & \textbf{0.836} (+19.6\%) & \textbf{0.858} (+32.6\%)
\\
\hline
\\[-6pt]
\end{tabular}
\end{center}
\vspace{-1em}
\end{table}

\begin{figure}[!htbp]
\centering
\begin{subfigure}{0.33\linewidth}
  \centering
  \includegraphics[width=1.3in]{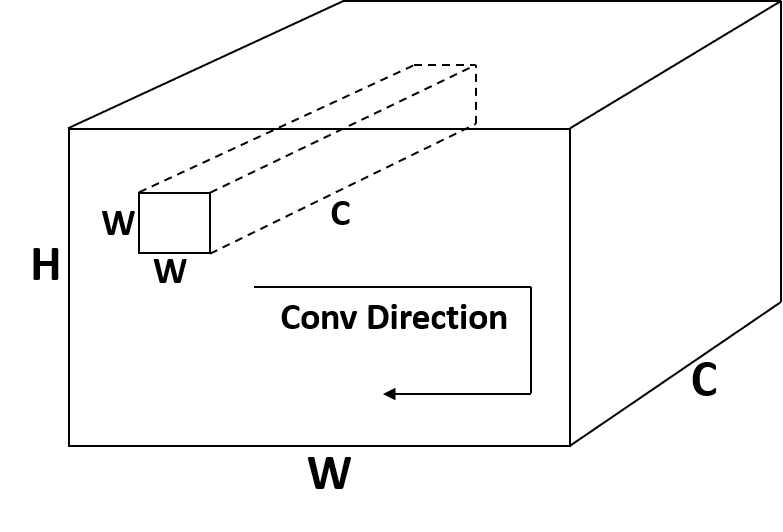}
  \caption{2D convolution}
  \label{fig:2d}
\end{subfigure}%
\hspace{0.15\linewidth}
\begin{subfigure}{0.33\linewidth}
  \centering
  \includegraphics[width=1.3in]{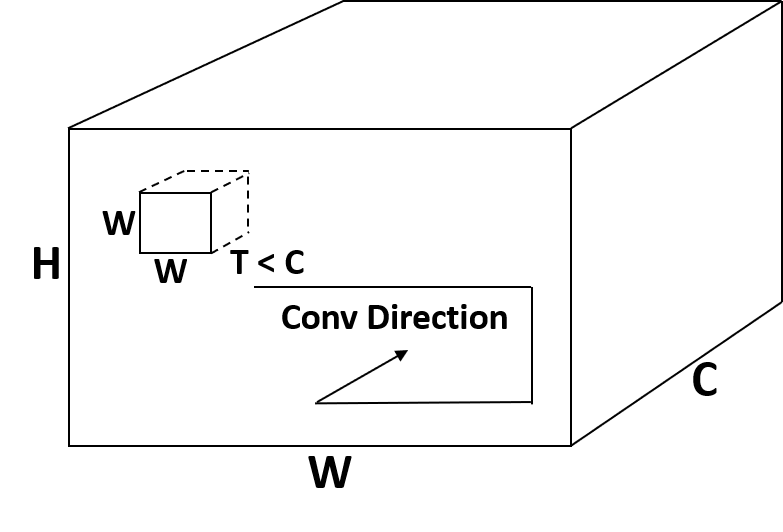}
  \caption{3D convolution}
  \label{fig:3d}
\end{subfigure}
\caption{Convolution Operation Comparison}
\label{fig:conv_operation}
\vspace{-1em}
\end{figure}

Table \ref{tab:2d_3d} shows the correlation score in both scenarios. The ``\begin{math} init \end{math}'' stands for the correlation score between the initial density and the label. We take the ``\begin{math} channel_{all} \end{math}'' and the ``\begin{math} channel_{seg} \end{math}'' to be two comparison models. The testing data is the same for both 3D convolution operation and 2D convolution operation. Therefore, the correlation scores between the initial density and the label are the same for both scenarios. Compared to the result in Section~\ref{sec:init}, without the training, the prediction accuracy drops for both models. However, from the results, we can notice that the performance drop with 2D convolution operation is less severe in comparison. The correlation score of the ``\begin{math} channel_{all} \end{math}'' model with 3D convolution operation is even lower than the ``\begin{math} init \end{math}''. The reason is that the features from the most recent time and early time periods are not distinguishable as the 3D convolution uses the same cube filter for all cycles. However, for the 2D convolution, each filter has a spatial extent. The number of spatial extents is equal to the number of input channels. The spatial extent increases the representability of the model. The demonstrations of the 2D and 3D convolution operations are given in Figure \ref{fig:conv_operation}. This suggests that using 2D convolution layers to extract features from each channel in our scenario is a more robust approach compared to using the 3D convolution. Consequently, we choose the 2D convolution operation to be the backbone of the feature extractor in our final model design.


The second observation from the result is that, the ``\begin{math} channel_{seg} \end{math}'' model always has a better prediction performance than the ``\begin{math} channel_{all} \end{math}'' model. With \begin{math} 2D \end{math} convolution operation, compared to the ``\begin{math} init \end{math}'', the ``\begin{math} channel_{seg} \end{math}'' model can achieve 19.6\% and 32.6\% correlation score improvement in NTB\_{30} and NTA\_{30}, respectively. However, the ``\begin{math} channel_{all} \end{math}'' model can only achieve 9.4\% and 16.1\% improvement. This result is consistent with our hypothesis at the beginning of the section. In our ``\begin{math} channel_{seg} \end{math}'' model, the missions with different launching time can be distinguished. Our model is capable of learning the meaningful features from the non-stationary missions.

\subsection{Density Prediction with No-Fly Zone Avoidance and Routing Algorithm}\label{sec:routing}

\begin{figure}[!htbp]
\centering
\begin{subfigure}{0.3\linewidth}
  \centering
  \includegraphics[width=\linewidth]{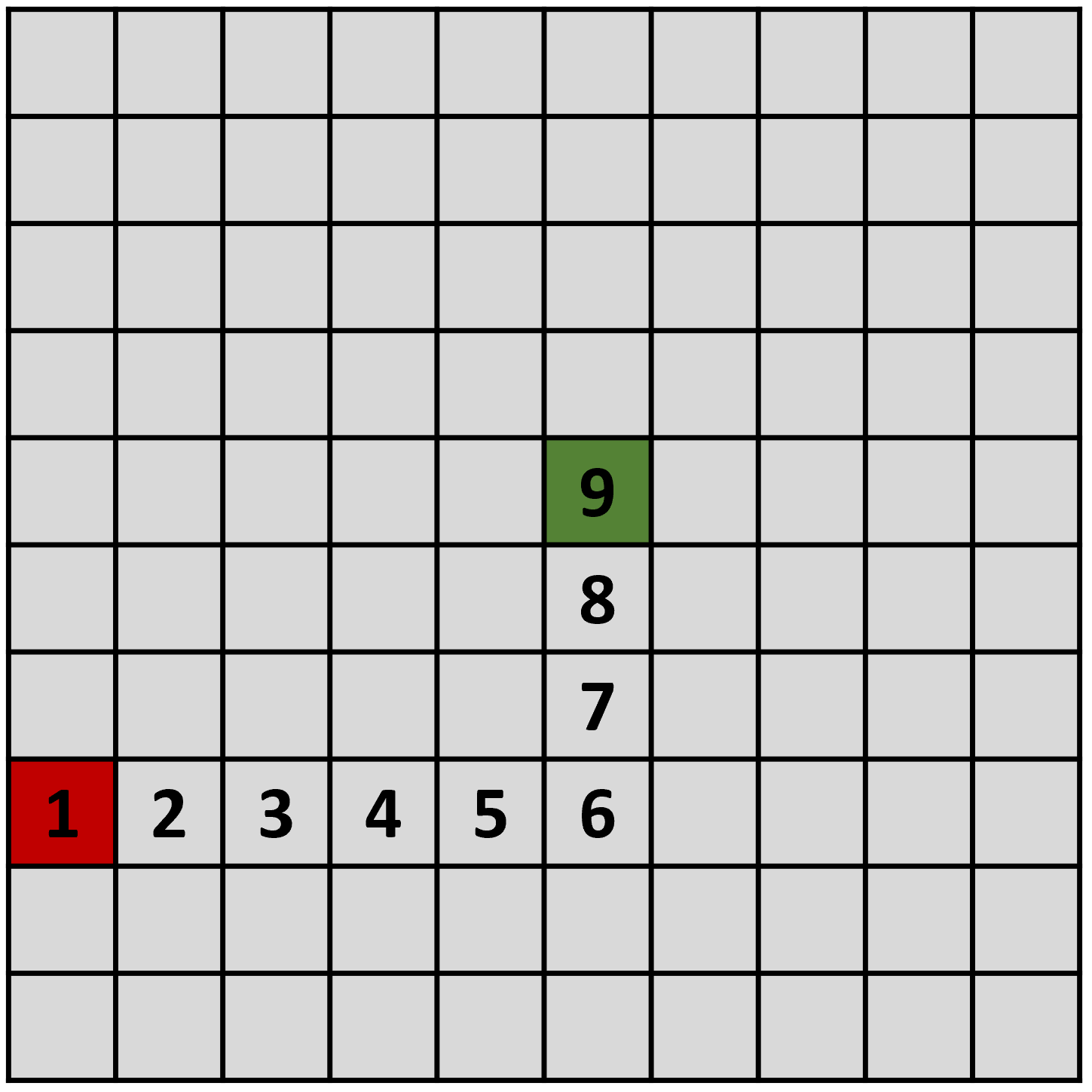}
  \caption{Flow}
  \label{fig:flow}
\end{subfigure}
\begin{subfigure}{0.3\linewidth}
  \centering
  \includegraphics[width=\linewidth]{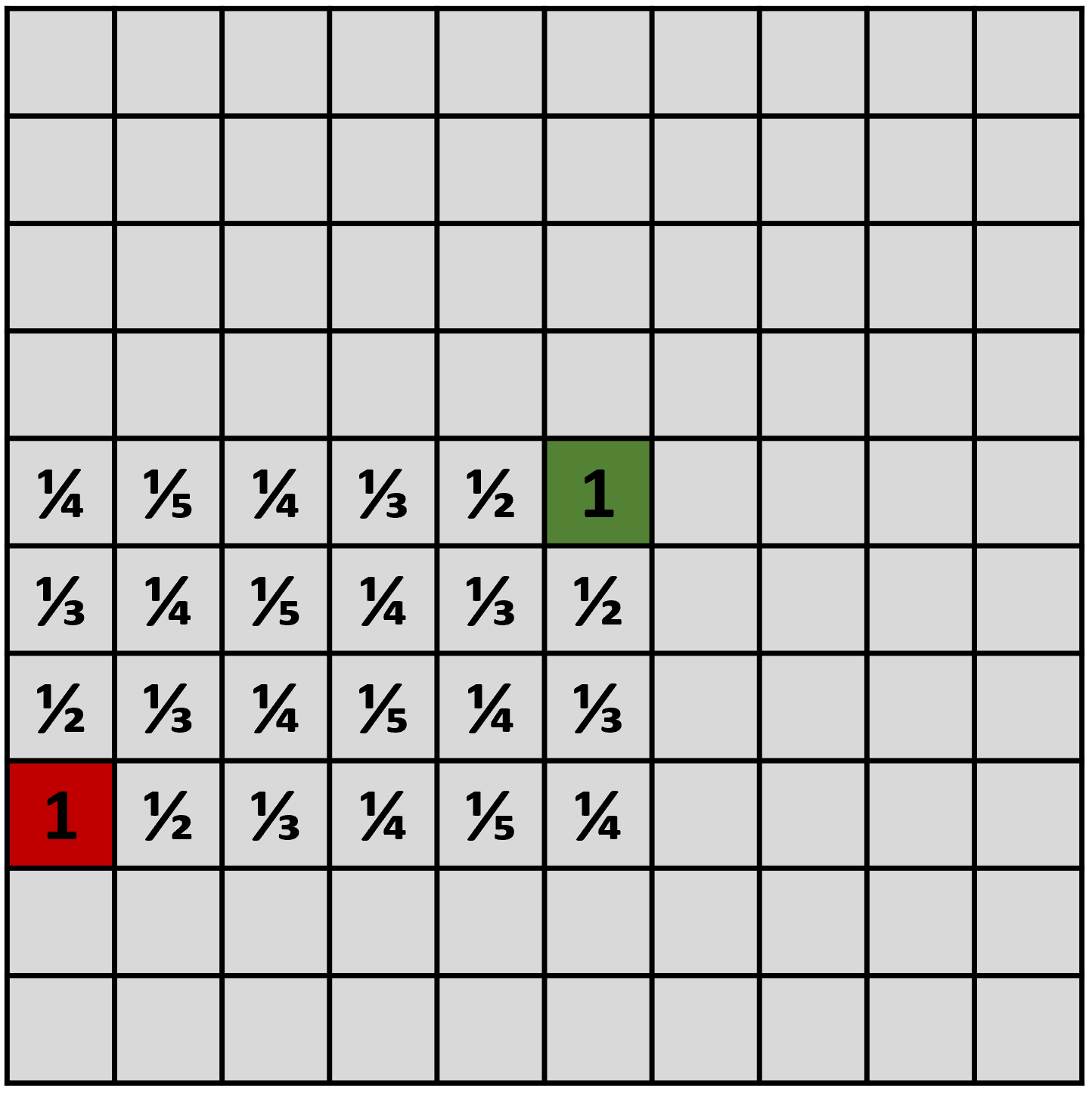}
  \caption{Probability}
  \label{fig:prob}
\end{subfigure}
\begin{subfigure}{0.3\linewidth}
  \centering
  \includegraphics[width=\linewidth]{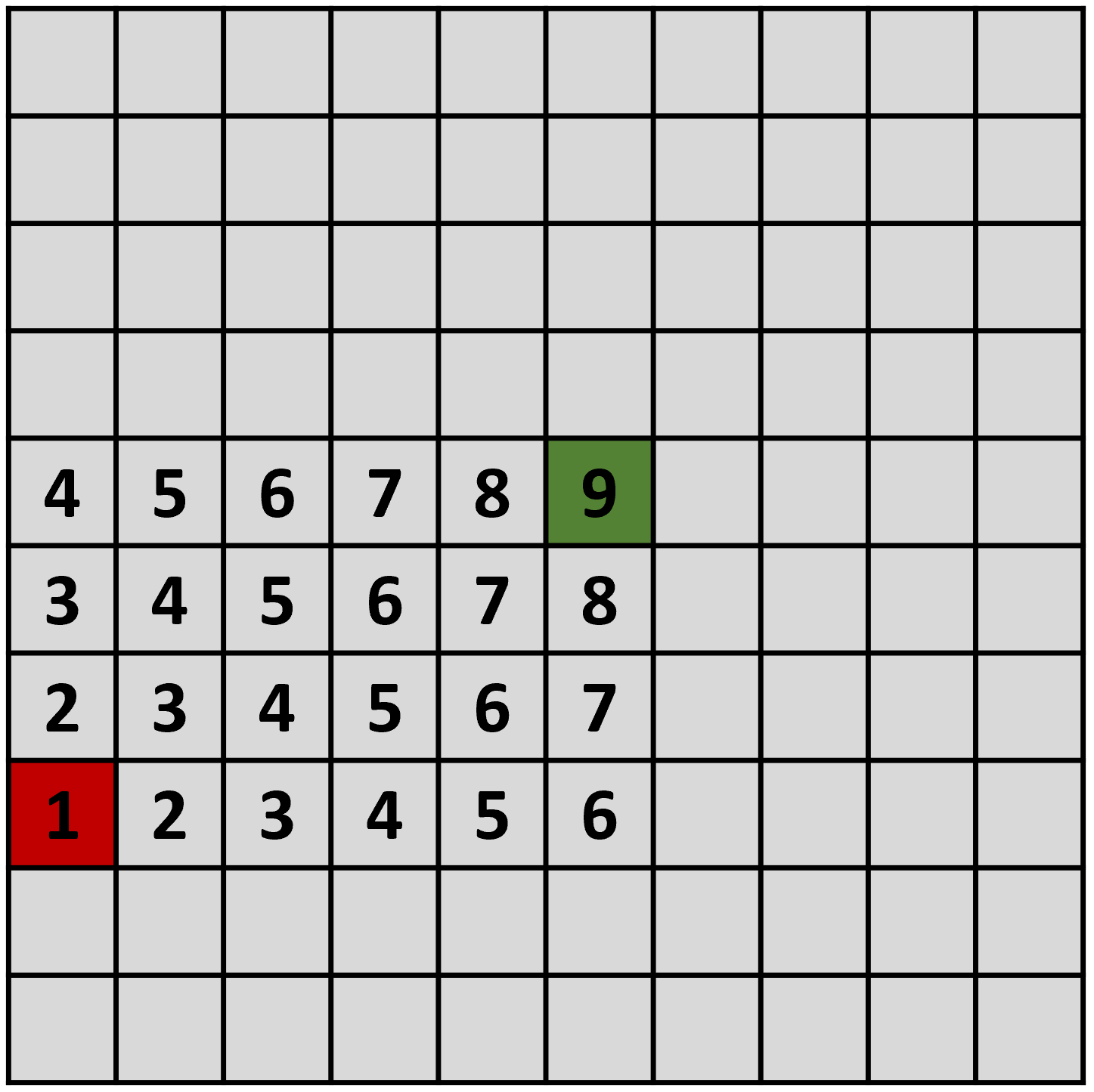}
  \caption{Ones}
  \label{fig:ones}
\end{subfigure}
\caption{Different Input Representation}
\label{fig:input}
\end{figure}

In the fourth experiment, we introduce no-fly zones into the simulation environment. In each batch of the simulation, the ratio of grids which are occupied by a no-fly zone is varying from 5\% to 45\%. By applying the routing algorithm, the simulated sUAS is capable of avoiding the no-fly zone and other sUASs. Hence, the sUAS trajectory is more heuristic and that leads to a more challenging density prediction task. In order to reach a high prediction accuracy, we investigate three potential input representations at the same time. The ``Flow'' input representation has been presented in Section~\ref{sec:mission}, as shown in Figure~\ref{fig:flow}. In the second input representation, we draw a bounding box to incorporate the launching and landing grids of each sUAS. The launching/landing grids are located at the two corners of the bounding box. The value in each grid represents the needed steps to move from the launching location, as shown in Figure~\ref{fig:ones}. Therefore, we call it ``Ones'' input representation. For the ``Probability'' input representation, we use the same bounding box to incorporate the launching and landing grids. However, the value in each grid stands for the probability that the sUAS moves from its previous adjacent grid to the current grid, as shown in Figure~\ref{fig:prob}.

\begin{table}[!htbp]
\begin{center}
\caption{Density Prediction with No-Fly Zone Avoidance and Routing Algorithm}\label{tab:routing}
\begin{tabular}{lcccc}
\hline
\rule{0pt}{12pt}
Correlation & Flow & Probability & Ones
\\
\hline
\\[-6pt]
\quad \begin{math} init \end{math} & 0.698 & 0.698 & 0.698\\
\quad \begin{math} channel_{all} \end{math} & 0.795 (+13.9\%) & 0.818 (+17.2\%) & \textbf{0.821} (+17.6\%)\\
\quad \begin{math} channel_{seg} \end{math} & 0.798 (+14.3\%) & 0.819 (+17.3\%) & \textbf{0.823} (+17.9\%)
\\
\hline
\\[-6pt]
\end{tabular}
\end{center}
\vspace{-1em}
\end{table}

Table \ref{tab:routing} shows the correlation scores for the different representations. From the result, we can notice that both models can improve the correlation in all the input representation types. However, the performance of the ``\begin{math} channel_{seg} \end{math}'' model is slightly better than all-channel model. Compared to the correlation score of the initial density, our presented ``\begin{math} channel_{seg} \end{math}'' model can improve the correlation score up to 14.33\% in "Flow" input representation, up to 17.34\% in "Probability" input representation and up to 17.91\% in the "Ones" input representation. The ``Ones'' input representation outperforms other two representations due to two reasons: 1) The routing algorithm is used in the simulation, therefore, the potential sUAS trajectories are more heuristic. Although the ``flow'' representation has the ability to indicate the sUAS movement, the flexibility of the model is also reduced by given only one possible path. 2) Compared to the ``Probability'' representation, the ``Ones'' representation does not only have all the possible trajectories, but also indicate the moving order of the sUAS. 

Besides the correlation evaluation, we also apply the AUROC to evaluate the performance of the ``\begin{math} channel_{seg} \end{math}'' model. In this experiment, the pixel whose value that is not zero in the label is considered to be the evaluation reference as we are more interested in the high dense area on the map. The \begin{math} P \end{math} estimation is a popular method in financial risk assessment and internet congestion investigation. Hence, we employ the \begin{math} P50 \end{math}, \begin{math} P75 \end{math}, \begin{math} P90 \end{math} and \begin{math} P99 \end{math} to select the threshold. After the threshold is defined, the pixel in the label whose value is larger than the threshold is binarized to 1, and vice versa. Table \ref{tab:p_threshold} shows the selected threshold for different \begin{math} P \end{math} values. For the prediction, the threshold value is sampled progressively from 0.0 to 1.0, with the 0.01 granularity.

\begin{table}[!htbp]
\begin{center}
\caption{Threshold Selection in Different P Value}\label{tab:p_threshold}
\begin{tabular}{lccccc}
\hline
\rule{0pt}{12pt}
 & P50 & P75 & P90 & P99
\\
\hline
\\[-6pt]
\quad \begin{math} Threshold  \end{math} & 0.2 & 0.3 & 0.5 & 0.8\\
\hline
\\[-6pt]
\end{tabular}
\end{center}
\vspace{-1em}
\end{table}

\begin{figure}[!htbp]
\centering

\begin{subfigure}{0.45\linewidth}
  \centering
  \includegraphics[width=\linewidth]{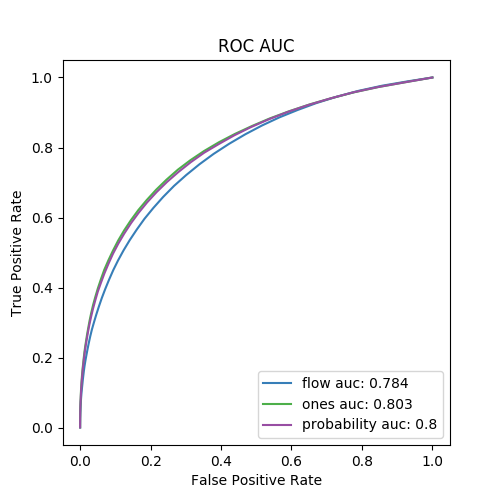}
  \caption{AUROC @ P50}
  \label{fig:traj_1_auroc}
\end{subfigure}
\begin{subfigure}{0.45\linewidth}
  \centering
  \includegraphics[width=\linewidth]{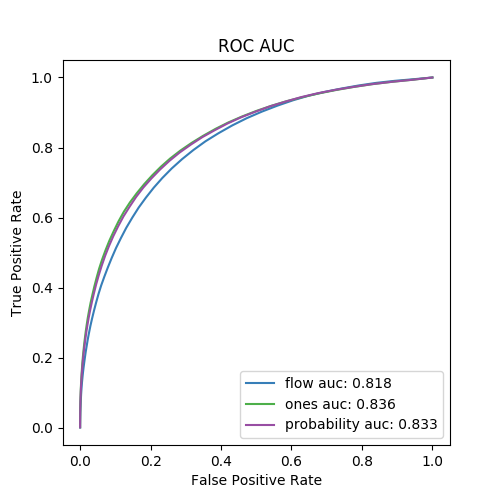}
  \caption{AUROC @ P75}
  \label{fig:traj_2_auroc}
\end{subfigure}

\begin{subfigure}{0.45\linewidth}
  \centering
  \includegraphics[width=\linewidth]{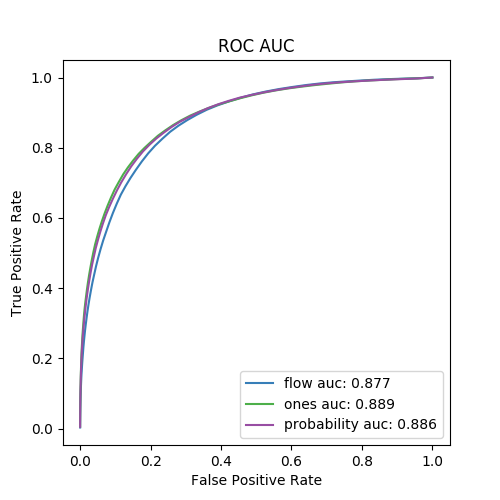}
  \caption{AUROC @ P90}
  \label{fig:traj_3_auroc}
\end{subfigure}
\begin{subfigure}{0.45\linewidth}
  \centering
  \includegraphics[width=\linewidth]{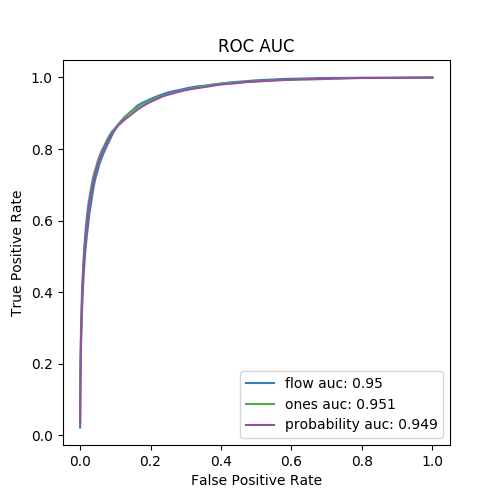}
  \caption{AUROC @ P99}
  \label{fig:traj_4_auroc}
\end{subfigure}

\caption{AUROC at Different P Value}
\label{fig:auroc}
\vspace{-1em}
\end{figure}

Figure \ref{fig:auroc} shows the AUROC in different thresholds. As we can see from the figure, the ``Ones'' input representation still outperforms other methods. The \begin{math} P50\end{math} means that half of the UAS flight areas are considered as the hot spot. In this strict circumstance, the AUROC of the ``Ones'' representation can still achieve 0.803. However, in reality, the severity is often exaggerated by choosing the \begin{math} P50\end{math}. For the \begin{math} P75, P90, and\ P99\end{math}, the AUROC of the ``Ones'' representation are 0.836, 0.889 and 0.951, respectively. This result further proves that our model is capable of making an accurate hot spot prediction.

\subsection{Density Prediction Visualization}\label{sec:visualization}

\begin{figure}[!htbp]
\centering

\begin{subfigure}{0.45\linewidth}
  \centering
  \includegraphics[width=\linewidth]{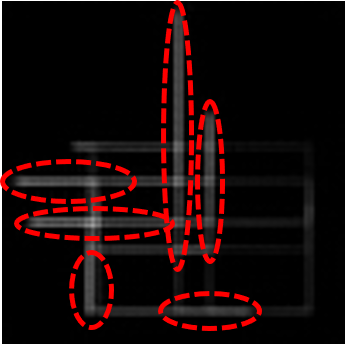}
  \caption{Density Prediction (Dense)}
  \label{fig:traj_1}
\end{subfigure}
\begin{subfigure}{0.45\linewidth}
  \centering
  \includegraphics[width=\linewidth]{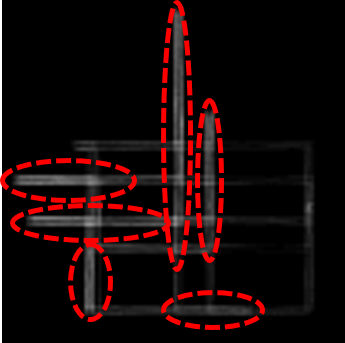}
  \caption{Density Label (Dense)}
  \label{fig:traj_2}
\end{subfigure}

\begin{subfigure}{0.45\linewidth}
  \centering
  \includegraphics[width=\linewidth]{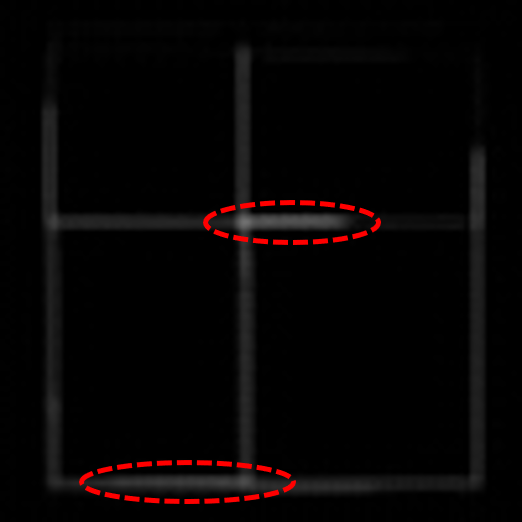}
  \caption{Density Prediction (Sparse)}
  \label{fig:traj_3}
\end{subfigure}
\begin{subfigure}{0.45\linewidth}
  \centering
  \includegraphics[width=\linewidth]{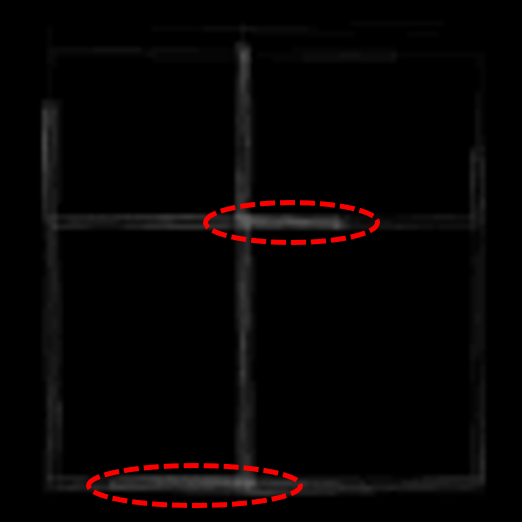}
  \caption{Density Label (Sparse)}
  \label{fig:traj_4}
\end{subfigure}

\caption{Density Prediction without Routing Visualization}
\label{fig:denst_normal}
\vspace{-1em}
\end{figure}

\begin{figure}[!htbp]
\centering

\begin{subfigure}{0.45\linewidth}
  \centering
  \includegraphics[width=\linewidth]{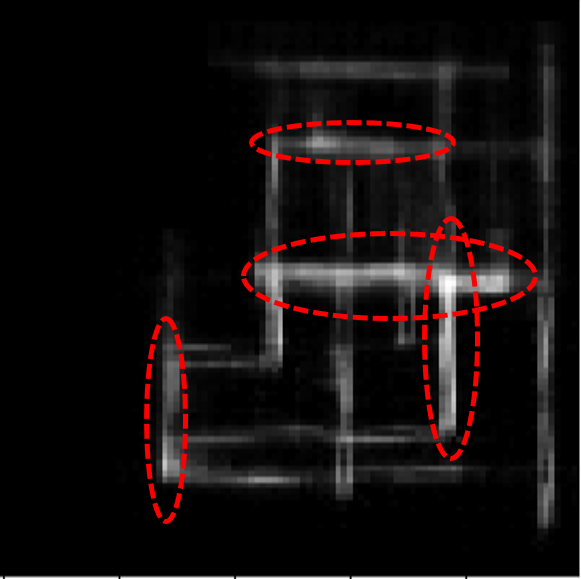}
  \caption{Density Prediction (Dense)}
  \label{fig:one_a}
\end{subfigure}
\begin{subfigure}{0.45\linewidth}
  \centering
  \includegraphics[width=\linewidth]{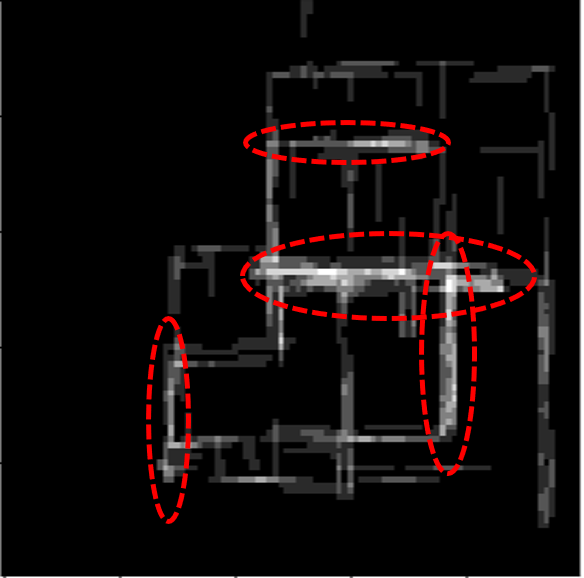}
  \caption{Density Label (Dense)}
  \label{fig:one_b}
\end{subfigure}

\begin{subfigure}{0.45\linewidth}
  \centering
  \includegraphics[width=\linewidth]{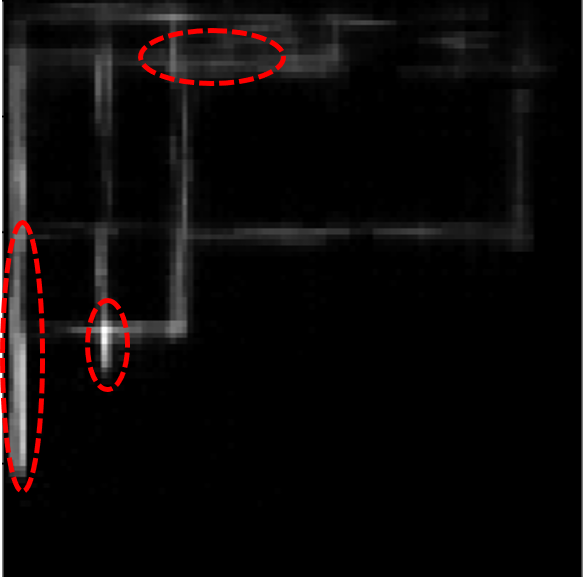}
  \caption{Density Prediction (Sparse)}
  \label{fig:one_c}
\end{subfigure}
\begin{subfigure}{0.45\linewidth}
  \centering
  \includegraphics[width=\linewidth]{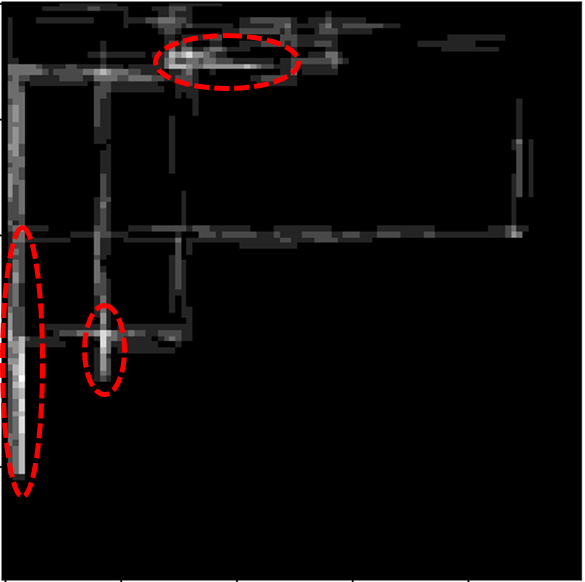}
  \caption{Density Label (Sparse)}
  \label{fig:one_d}
\end{subfigure}

\caption{Density Prediction with Routing Visualization}
\label{fig:denst_routing}
\vspace{-1em}
\end{figure}

In the last experiment, we visualize several UAS density predictions to give a qualitative demonstration of our model. We select two typical scenarios to test the performance: a) dense traffic, b) sparse traffic. Both scenarios are tested with/without the routing algorithm. Figure~\ref{fig:denst_normal} shows the prediction results without the routing algorithm. The left side figures are the prediction and the right side figures are the label. The value of each pixel varies from 0 to 1, representing the average density in \begin{math} T_1 \end{math}. The brighter area means that there are more sUAS passing through this location. The density prediction with the routing algorithm is shown in Figure~\ref{fig:denst_routing}.

From Figure~\ref{fig:denst_normal} we can notice that, our predicted densities are close to the label in both scenarios. In the dense traffic scenario, all 6 dense areas are predicted by our model, as shown in Figure~\ref{fig:traj_1} and Figure~\ref{fig:traj_2}. Although some of the dense areas are close with each other, the model can still predict them clearly. In the sparse traffic scenario, there are only two horizontal dense areas. One is in the middle of the map, the other is at the bottom of the map. Both of them are predicted accurately by our model.

After the routing algorithm is introduced, the sUAS trajectory is heuristic which will lead to a more random density distribution. Both prediction and label become blurry in this situation. Under this circumstance, our presented model can still predict the most obvious dense areas. In Figure~\ref{fig:one_b}, there are four obvious dense areas which are marked by the red dash circle. Figure~\ref{fig:one_a} shows that all of the dense areas are predicted successfully by our model. In the sparse traffic scenario, there are three obvious dense areas in Figure~\ref{fig:one_d}. Although the model fails to predict the dense area at the top of the map, two other dense areas at the left bottom have been predicted successfully, as shown in Figure~\ref{fig:one_c}.

\section{Conclusions} \label{sec:conclusion}
In this paper, we have proposed a novel mission-aware spatio-temporal model, which aims at predicting the UAS instantaneous density. The model has the ability to extract meaningful features from the given historical density and learn the information from the pre-scheduled missions. Compared to the baseline models, for simplified traffic scenario where no-fly zones and safe distance among sUASs are not considered, our model improves the prediction accuracy by more than 15.2\% and its correlation score reaches 0.947. The results in Section \ref{sec:2d_3d} show that our model is sensitive to the pre-scheduled missions and has the ability to predict the transient behavior of the traffic distribution. In a more realistic scenario, where the no-fly zone avoidance and the safe distance among sUASs are maintained using A* routing algorithm, our model can still achieve a correlation score of 0.823. Moreover, the AUROC results demonstrate that the hot spot predicted by our model is accurate. The qualitative results also show that the presented model can generate a detailed prediction.


\bibliographystyle{IEEEtran}
\bibliography{reference}
%





\end{document}